%% file: main.tex
\documentclass[runningheads]{llncs}

\usepackage{eccv}

\usepackage{eccvabbrv}

\usepackage{graphicx}
\usepackage{booktabs}

\usepackage[accsupp]{axessibility}  

\usepackage[breaklinks,colorlinks]{hyperref}

\usepackage{orcidlink}

\usepackage{multirow}
\definecolor{dark-gray}{gray}{0.20}
\definecolor{middle-gray}{gray}{0.85}
\definecolor{light-gray}{gray}{0.93}
\definecolor{mygreen}{HTML}{39b54a}
\definecolor{myred}{HTML}{ff5050}
\newcommand{\inchl}[2]{
	#1 \fontsize{7.5pt}{1em}\selectfont\color{mygreen}{$\uparrow$ #2}
}
\newcommand{\bfinchl}[2]{
	\textbf{#1} \fontsize{7.5pt}{1em}\selectfont\color{mygreen}{$\uparrow$ \textbf{#2}}
}

\newcommand{\pub}[1]{{\color{dark-gray}{\tiny{[{#1}]}}}}

\usepackage{colortbl}

\input{math_commands}

\begin{document}

\title{Online Vectorized HD Map Construction using Geometry} 


\author{Zhixin Zhang\inst{1}\orcidlink{0000-0003-3107-0004}
Yiyuan Zhang\inst{2}\orcidlink{0000-0001-6643-9698} 
Xiaohan Ding\inst{3}  
Fusheng Jin\inst{1}\thanks{Corresponding author}
Xiangyu Yue\inst{2}
}

\authorrunning{Z.~Zhang et al.}

\institute{
\textsuperscript{1}School of Computer Science and Technology, Beijing Institute of Technology \\
\textsuperscript{2} The Chinese University of Hong Kong
\quad
\textsuperscript{3} Tencent AI Lab~~~ \\
\texttt{zhangzhixin@bit.edu.cn, \quad yiyuanzhang.ai@gmail.com} 
\\
\url{https://invictus717.github.io/GeMap/}
}

\maketitle

\input{sec/0_abstract}
\input{sec/1_intro}
\input{sec/2_related}
\input{sec/3_method}
\input{sec/4_experiment}
\input{sec/5_conclusion}

\noindent \textbf{Acknowledgments.}
This work is partially supported by the National Natural Science Foundation of China (No. 62272045, No. 8326014), The Shun Hing Institute of Advanced Engineering (No. 8115074), and CUHK Direct Grants (No. 4055190).

\clearpage  

%
%
\bibliographystyle{splncs04}
\bibliography{egbib}

\input{sec/X_suppl}

\end{document}

%% file: math_commands.tex
\usepackage{amsmath}
\usepackage{bm}

\def\eqref#1{equation~\ref{#1}}

\def\1{\bm{1}}

\def\vp{{\bm{p}}}

\def\vv{{\bm{v}}}

\def\vy{{\bm{y}}}

\def\mE{{\bm{E}}}

\def\mI{{\bm{I}}}

\def\mL{{\bm{L}}}
\def\mM{{\bm{M}}}

\def\mP{{\bm{P}}}

\def\mW{{\bm{W}}}

\def\mY{{\bm{Y}}}

\DeclareMathAlphabet{\mathsfit}{\encodingdefault}{\sfdefault}{m}{sl}
\SetMathAlphabet{\mathsfit}{bold}{\encodingdefault}{\sfdefault}{bx}{n}

\def\sI{{\mathbb{I}}}

\def\sM{{\mathbb{M}}}

\newcommand{\R}{\mathbb{R}}

\newcommand{\normlone}{L^1}


%% file: sec/0_abstract.tex
\begin{abstract}
Online vectorized High-Definition (HD) map construction is critical for downstream prediction and planning.
Recent efforts have built strong baselines for this task, however, geometric shapes and relations of instances in road systems are still under-explored, such as parallelism, perpendicular, rectangle-shape, \etc. In our work, we propose GeMap (\textbf{Ge}ometry \textbf{Map}), which end-to-end learns Euclidean shapes and relations of map instances beyond fundamental perception. Specifically, we design a geometric loss based on angle and magnitude clues, robust to rigid transformations of driving scenarios. To address the limitations of the vanilla attention mechanism in learning geometry, we propose to decouple self-attention to handle Euclidean shapes and relations independently. GeMap achieves new state-of-the-art performance on the nuScenes and Argoverse 2 datasets. Remarkably, it reaches a 71.8\% mAP on the large-scale Argoverse 2 dataset, outperforming MapTRv2 by +4.4\% and surpassing the 70\% mAP threshold for the first time. Code is available at \url{https://github.com/cnzzx/GeMap}.
\keywords{HD Map Construction \and Geometry Representation \and Geometry-Decoupled Attention}
\end{abstract}

%% file: sec/1_intro.tex
\section{Introduction}
Vectorized HD maps provide structured environmental information for autonomous vehicles and have been widely adopted in downstream tasks, such as trajectory forecasting~\cite{liang2020learning, deo2022multimodal, zhou2022hivt} and planning~\cite{scheel2022urban, espinoza2022deep}. Online Vectorized HD Map Construction can significantly reduce the need for labor-intensive annotations and facilitate real-time updates in autonomous driving~\cite{liao2023maptr,ding2023pivotnet,li2022hdmapnet}.

\begin{figure}[t]
	\centering
        \subfloat[ 
    	\label{fig:inv-a}
        ]{
	\begin{minipage}{0.43\linewidth}{
			\begin{center}
				{\includegraphics[width=1\linewidth]{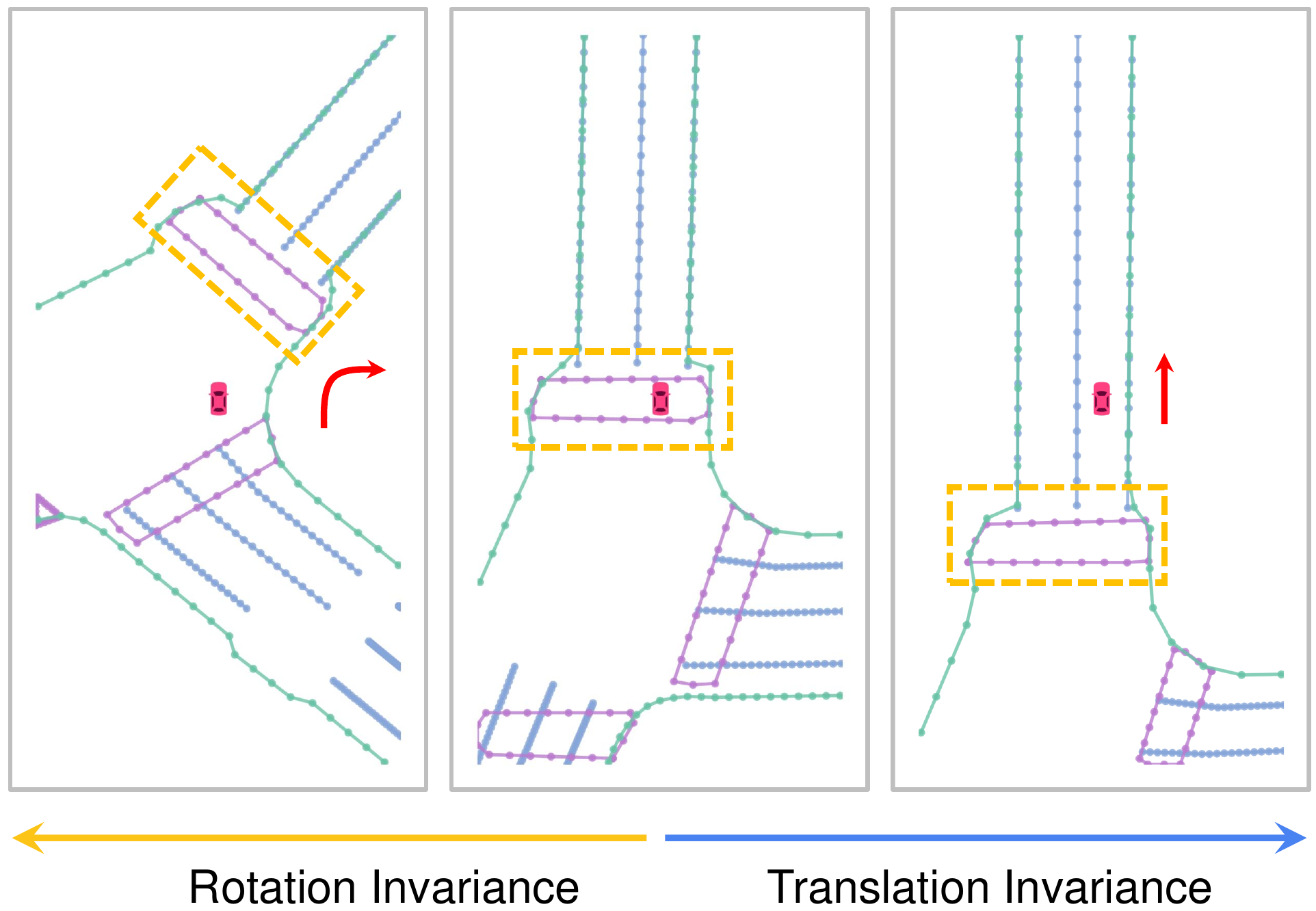}}
			\end{center}
		}
	\end{minipage}
        }
        \subfloat[
            \label{fig:inv-b}
        ]{
        \begin{minipage}{0.48\linewidth}{
			\begin{center}
				{\includegraphics[width=1\linewidth]{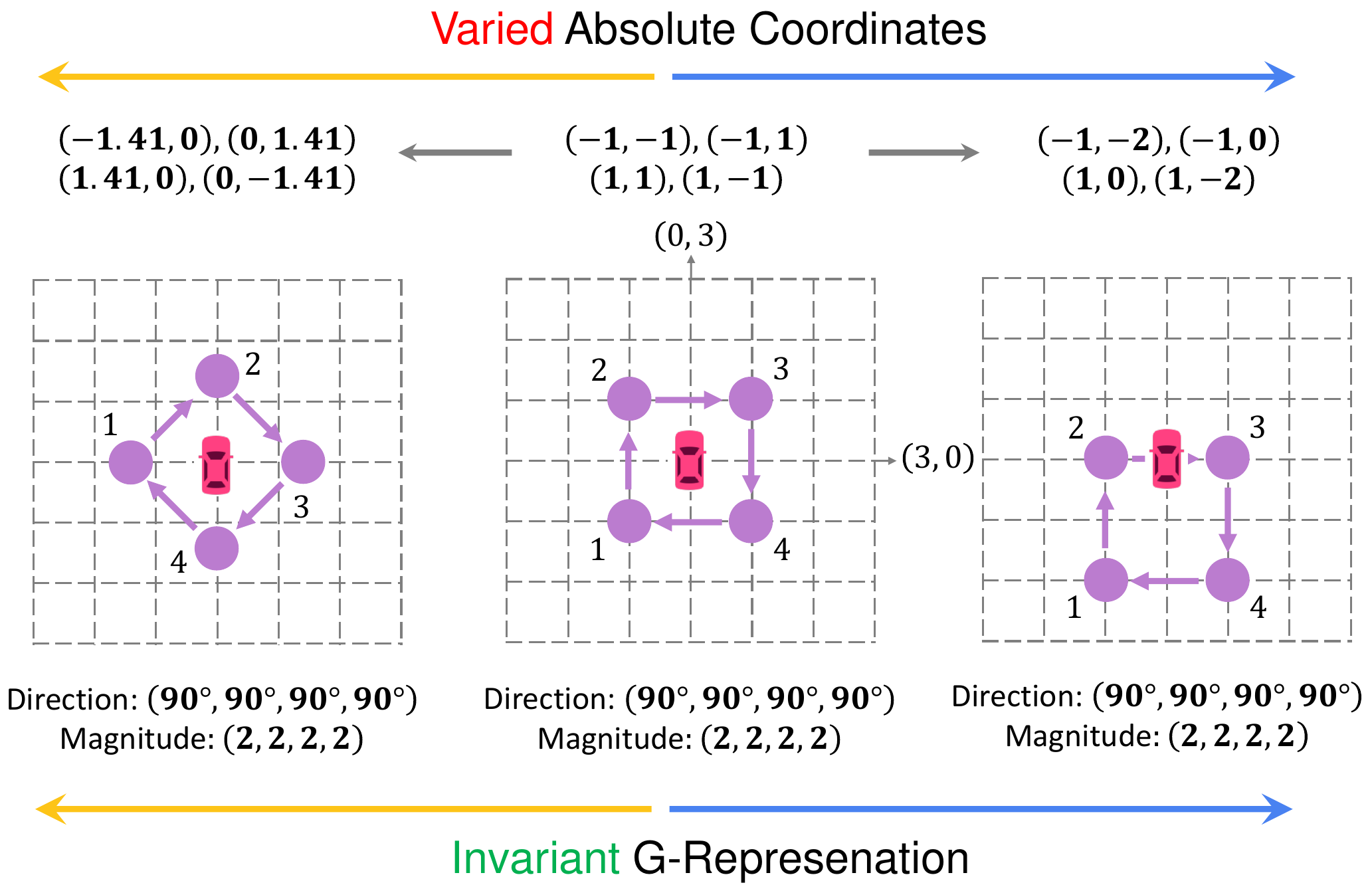}}
			\end{center}
		}
	\end{minipage}
        }
    \vspace{-2mm}
	\caption{ \textbf{Geometric Invariance}. (a) As the ego vehicle moves, after rotation ($\leftarrow$) and translation ($\rightarrow$), the shape of the crossing and the parallelism between lanes remain unchanged, which indicates the invariant property of geometry to rigid transformations. (b) Absolute coordinates are vulnerable to rotation and translation, however, our G-representation is invariant, which is more suitable to capture geometric properties.}
    \vspace{-3.5mm}
	\label{fig:inv}
\end{figure}
With the development of Bird's-Eye-View (BEV) representation, online HD map construction has achieved significant advancements~\cite{liao2023maptrv2,li2022bevformer,gu2023vip3d}.
Early works~\cite{jiao2018machine, lu2019monocular, philion2020lift, yang2021projecting} formulate HD map construction as a dense prediction task. However, these methods generate maps in image format, which is redundant for representing sparse map instances. Then, a more compact map formulation was introduced to minimize redundancy, albeit at the cost of adding time-consuming post-processing steps~\cite{li2022hdmapnet}.
In response to this, recent works~\cite{liu2023vectormapnet, liao2023maptr, qiao2023end, ding2023pivotnet} attempt to end-to-end construct vectorized HD maps to avoid extensive post-processing. These methods typically sample points from map instances and represent each instance as a polyline~\cite{liu2023vectormapnet, liao2023maptr, ding2023pivotnet} or parameterized curve~\cite{qiao2023end}.

\begin{figure}[tb]
    \centering
    \subfloat[ 
	\label{fig:idea-a}
    ]{
	\begin{minipage}{0.88\linewidth}
            \begin{center}
                \includegraphics[width=1\linewidth]{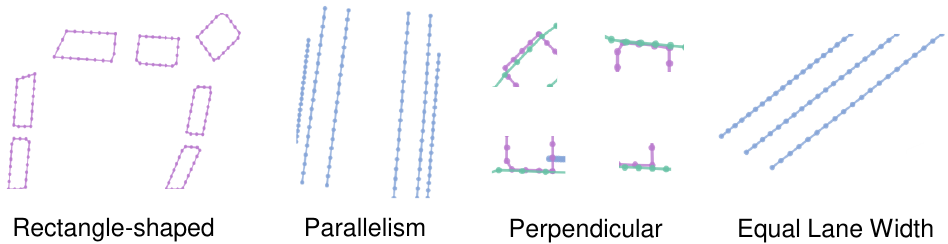}
    	\end{center}
        \end{minipage}
    } \\
    \subfloat[ 
	\label{fig:idea-b}
    ]{
        \begin{minipage}{0.44\linewidth}
            \begin{center}
                \includegraphics[width=1\linewidth]{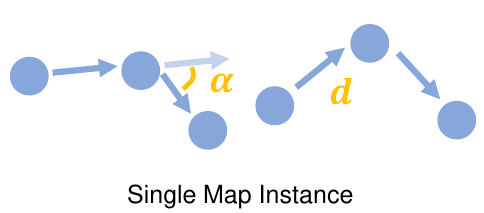}
            \end{center}
        \end{minipage}
    }
    \subfloat[ 
	\label{fig:idea-c}
    ]{
	\begin{minipage}{0.44\linewidth}
            \begin{center}
			\includegraphics[width=1\linewidth]{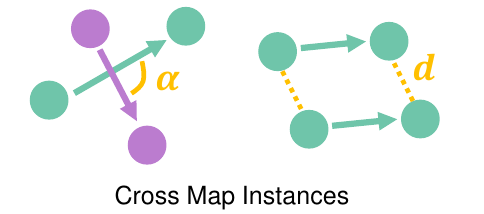}
	    \end{center}
        \end{minipage}
    } 
    \vspace{-2mm}
    \caption{
        \textbf{Geometric properties and G-Representation.} (a) Geometry in the transportation road system. (b) (c) We propose to model the geometric properties of a single map instance and multiple instances, with magnitude $d$ and angle $\alpha$.
    }
    \vspace{-3.5mm}
\label{fig:idea}
\end{figure}

We observe that transportation road systems exhibit significant geometric characteristics (Figure~\ref{fig:idea-a}), such as parallel lanes, perpendicular crossings, equal lane widths, \etc.
However, these geometric properties of shapes and relations between map instances have not been fully explored. Meanwhile, there are two notable limitations among existing methods~\cite{liu2023vectormapnet, liao2023maptr, qiao2023end, ding2023pivotnet} which can be alleviated by leveraging these geometric properties. \textbf{1)} \textit{Excessive dependency on absolute coordinates}: as the ego-vehicle moves, instances experience rotations and translations, as depicted in Figure~\ref{fig:inv-a}. However, widely adopted representations such as polylines and parameterized curves ~\cite {gao2020vectornet, qiao2023end} are inherently sensitive to rotation and translation changes. \textbf{2)} \textit{Objective conflicts of vanilla attention mechanism (detail in \S~\ref{sec:dattn})} struggles to learn diverse shapes and relations, even though geometric properties such as rectangle shape, parallelism, and perpendicular relationships are commonly found in driving scenarios, as illustrated in Figure~\ref{fig:idea-a}. We believe that incorporating these geometric properties significantly enhances the precision and efficiency of online HD map construction.

To address the above two limitations, we introduce a novel geometric representation that captures shapes of individual map instances and relations between different instances as illustrated in Figure~\ref{fig:idea-b} and \ref{fig:idea-c}, referred to as \textbf{G-Representation}. It enhances the vanilla representation of map instances by incorporating a translation- and rotation-invariant representation that effectively leverages instance geometry. The local structures of map features are encoded using \textit{displacement vectors}. These vectors are computed from the absolute coordinates of polyline points to effectively represent the relative positions and orientations of adjacent points. To quantify these features within Euclidean space, we employ both the \textit{magnitudes} of the displacement vectors and the \textit{angles} formed between these vectors. Based on G-Representation, we propose the \textbf{Euclidean Shape Clues} to represent the shapes of individual map instances and the \textbf{Euclidean Relation Clues} to represent the inter-instance relations in Euclidean space.

G-Representation is a simple yet effective method to address the abovementioned two limitations: \textbf{1)} \textit{Geometric Invariance}, by concentrating on the relative relationships between points within and across map instances, it inherently attains translation and rotation invariance as illustrated in Figure~\ref{fig:inv-b}. This enhances its robustness against variations in data collection and equips it to effectively handle different coordinate systems. \textbf{2)} \textit{Euclidean Modeling}, it captures the inherent geometry in transportation road systems. Diverse shapes and relation geometry can be simplified as enumerations of magnitudes and directions. For instance, in Euclidean space, parallel lanes can be easily modeled as approximate directions (${\alpha\approx0}$).

Building upon G-Representation, we propose a framework named \textbf{GeMap} for HD map construction. A BEV encoder is used to extract features from multi-view input images, while a geometry-decoupled decoder is employed to focus on geometric aspects.   
Specifically, we adapt the attention mechanism~\cite{vaswani2017attention} and propose \textbf{Geometry-Decoupled Attention (GDA)}. GDA sequentially applies attention to queries belonging to the same instance and attention to queries across different instances. This can significantly boost the geometry learning of key points of shapes and relations between map instances. 
Furthermore, we propose an objective function named \textit{Euclidean Loss} to optimize G-Representation. {Specifically, we transform the conventional polylines of the ground truth map into our G-Representation. In this way, the model gets optimized to better understand the magnitude of displacement vectors and the angles between them, thereby facilitating a more effective learning of geometric properties.}

Experiments on nuScenes and Argoverse 2 datasets demonstrate the effectiveness of GeMap. We reach new state-of-the-art performances on both datasets. With camera images only, GeMap achieves 69.4\% and 71.8\% mAP on the nuScenes and Argovserse 2, respectively. Visualization results (\S~\ref{sec:vis}) further demonstrate the better perception of shape and relation, alleviation of occlusion, and robustness to rigid transformations of GeMap.

Our contributions are summarized as the following:

\begin{itemize}

\item  We propose {G-Representation} which harnesses critical geometric properties of rotation and translation invariance in autonomous driving scenarios, opening up new research avenues within the field.

\item We introduce GeMap, a novel framework incorporating geometry-decoupled attention and Euclidean Loss function, specifically designed to learn the intrinsic geometry of online HD maps.

\item GeMap achieves new state-of-the-art results in HD map construction on the nuScenes and Argoverse 2 datasets, notably surpassing the 70\% mAP on the large-scale Argoverse 2 dataset for the first time.
\end{itemize}

%% file: sec/2_related.tex
\section{Related Work}

\subsection{Online HD Map Construction}

Traditionally, HD map construction has required labor-intensive manual or semi-automatic annotations~\cite{jiao2018machine, mi2021hdmapgen}. To streamline this, recent studies~\cite{pan2020cross, philion2020lift, loukkal2021driving, zhou2022cross, li2022bevformer} have focused on online construction, approaching HD maps as a dense prediction challenge. Innovations such as MetaBEV~\cite{ge2023metabev} aim to mitigate sensor issues, while MVNet~\cite{xie2023mv} uses historical data for improved semantic consistency. The trend towards automatic vectorization of HD maps is spearheaded by works such as HDMapNet~\cite{li2022hdmapnet}, which fuses camera and LiDAR inputs in BEV space and is advanced by end-to-end solutions such as VectorMapNet~\cite{liu2023vectormapnet} and MapTR~\cite{liao2023maptr}, leveraging Transformer-based models. StreamMapNet~\cite{yuan2023streammapnet} and PivotNet~\cite{ding2023pivotnet} build upon this, modifying attention mechanisms for better performance. Our contribution starts from the strengths of the Transformer architecture, optimizing it with a decoupled self-attention block for enhanced geometric processing.

\subsection{Cross-view BEV Learning}

The conversion of Perspective View (PV) camera images into a unified BEV space is a significant challenge for autonomous driving systems. Previous studies such as~\cite{philion2020lift, li2023bevdepth} use depth estimation from monocular images for this transformation, others~\cite{chen2022efficient, li2022bevformer, liu2023bevfusion} have developed PV-to-BEV conversion methods without explicit depths. For example, GKT~\cite{chen2022efficient} employs a geometric-guided kernel, and BEVFormer~\cite{li2022bevformer} uses deformable attention for the conversion. However, these methods can distort shapes and geometric relations, leading to inaccuracies. Our approach integrates geometric supervision in prediction, addressing these geometric inconsistencies while acknowledging the value of depth information.

\subsection{Geometric Instance Modeling}
HD maps depict instances with diverse geometric properties such as pedestrian crossings and lane boundaries, which present vectorization challenges. Traditional vectorization approaches model these instances as polylines~\cite{tabelini2021keep, wang2022keypoint, liu2023vectormapnet, liao2023maptr} or polynomial curves~\cite{van2019end, tabelini2021polylanenet, feng2022rethinking, qiao2023end}, with some incorporating Bézier Curves for improved fitting~\cite{feng2022rethinking, qiao2023end}. However, such methods often overlook geometric properties including shapes, parallelism, perpendicular, and \etc. 
In 3D Lane Detection, some works~\cite{liu2022learning,huang2023anchor3dlane,li2022reconstruct} also explore simple geometry priors of lanes such as ``equal lane width''. However, they are not generalizable to more complex HD map instances. Existing works on modeling geometry properties of map instances mainly address rasterization-based shape loss~\cite{van2019end, zhang2024online} or edge loss~\cite{liao2023maptr}, and they fail to preserve cross-instance relationships. PivotNet~\cite{ding2023pivotnet} offers an architectural advancement with its line-aware point decoder, yet it still does not fully address the complexity of instance geometry. 

Given the current under-exploration of diverse geometric properties in shape and relation, we propose utilizing Euclidean shape and relation losses and decoupling the traditional self-attention module to empower the model with a more robust understanding of instance geometry.

%% file: sec/3_method.tex
\section{Method}

\subsection{Preliminary}\label{sec:form}

\begin{figure*}[ht]
	\centering
        \vspace{-3mm}
	\includegraphics[width=0.87\linewidth]{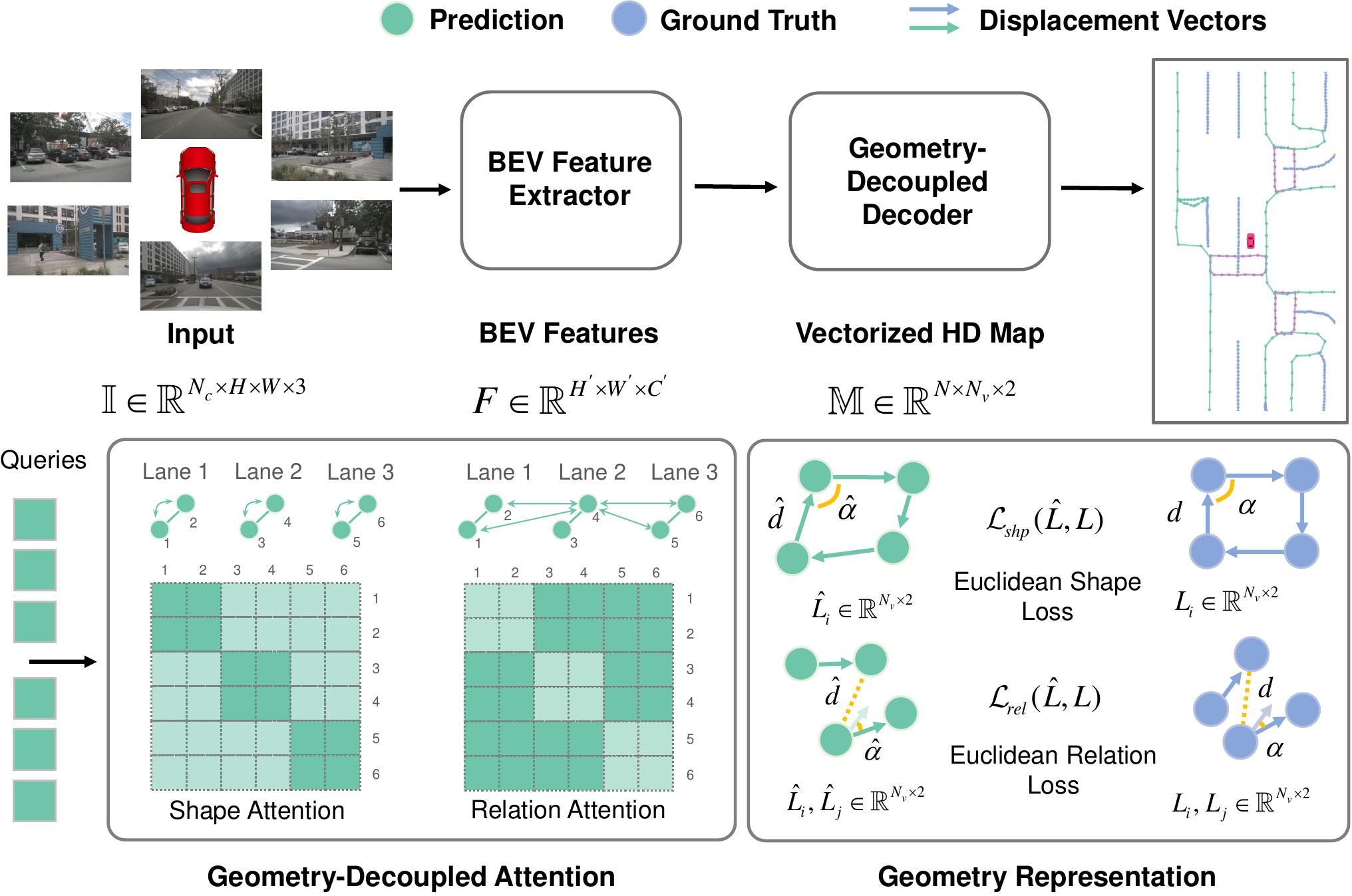}
        \vspace{-2mm}
	\caption{
		Illustration of our framework. First, PV images are transformed into BEV features, then a Geometry-Decoupled Decoder outputs the vectorized HD Map. In each block of the decoder, queries are first processed by Euclidean shape and relation attention, which focuses on geometric relevance. Finally, predictions are enhanced in G-Representations by shape and relation constraint.
	}
        \vspace{-3.5mm}
	\label{fig:framework}
\end{figure*}

The architecture of our method is illustrated in Figure~\ref{fig:framework}. Input images are denoted by $\sI = \{\mI_i\}_{i=1}^{N_c}$, where $\mI_i \in \R^{H\times W\times 3}$ and $N_c$ is the number of cameras. The output is a set of $N$ map instances $\sM= \{\mL_i\}_{i=1}^{N}$, where each map instance is represented by a polyline $\mL_i \in \R^{N_v\times 2}$, \ie, an ordered sequence of $N_v$ two-dimensional points. As shown in Figure~\ref{fig:framework}, the commonly adopted pipeline comprises three steps. \textbf{1)} A BEV feature extractor processes the multi-view images. \textbf{2)} The extracted BEV features are fed into a map decoder, which predicts the map instances based on the BEV features. \textbf{3)} To optimize the pipeline, the predicted map instances are compared against the ground truth. Beyond point-to-point comparison, G-representations of predictions and the ground truth are computed and the difference is measured by $\normlone$ loss. 

\subsection{Architecture Overview} \label{sec:arch}
\noindent\textbf{BEV Feature Extractor.} A shared vision backbone in different views and parameterized PV-to-BEV transformation network are employed to aggregate features from various perspectives. In our default configuration, we utilize ResNet~\cite{he2016deep} as the PV backbone and GKT~\cite{chen2022efficient} as the PV-to-BEV transformation network.

\noindent\textbf{Geometry-Decoupled Decoder.} We adopt a Transformer decoder to predict polylines. The decoder obtains $N\times N_v$ queries that represent points and processes them via self-attention and aggregate BEV features via deformable cross-attention~\cite{zhu2020deformable}. Finally, a prediction head is used to convert queries to polylines $\{\hat{\mL}_i\}_{i=1}^{N}$. We utilize a multi-layer perceptron as the polyline prediction head. 

\noindent\textbf{Geometric Loss.} For training, we convert both the predicted polylines and ground truth into the proposed G-Representation. Based on that, we let the model optimize the predicted magnitudes of displacement vectors and angles between displacement vectors to match the ground truth. 

In the following subsections, we first introduce Geometric Representation in \S~\ref{sec:local_geo} since it is the core of our framework. Based on that, we introduce Geometric Loss in \S~\ref{sec:loss} and Geometry-Decoupled Decoder in \S~\ref{sec:dattn}, respectively.

\subsection{Geometric Representation} \label{sec:local_geo}

\subsubsection{Euclidean Shape Clues.} \label{sec:shape_clue}
We first introduce the representation of shapes of individual map instances. For each instance, we describe the local geometry with displacement vectors between neighboring points which are computed as:
\begin{equation}
	\vv_{u}^{i} = \mL_{i,u+1} - \mL_{i,u}\quad(u \in \{1, 2, ..., N_v\}) \,,
\end{equation}
where we define $\mL_{i,N_v+1} := \mL_{i,1}$ to unify the geometric formulation of closed and open polylines.

\begin{figure}[htb]
	\centering
	\begin{minipage}{0.85\linewidth}{
        \begin{center}
            \includegraphics[width=1\linewidth]{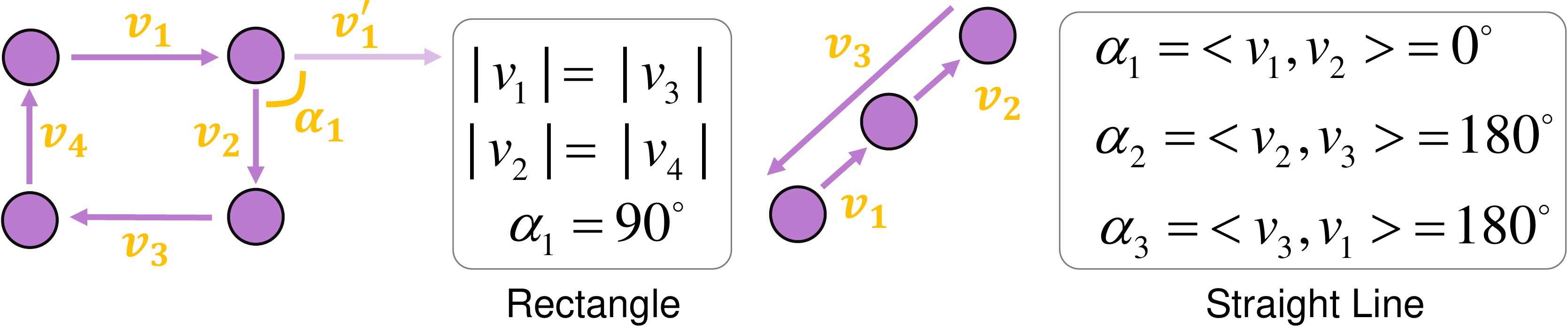}
        \end{center}}
        \end{minipage}
        \vspace{-2mm}
	\caption{
		Euclidean Shape Clues. Magnitudes of displacement vectors and angles between neighboring vectors indicate shape clues and are utilized to compute shape loss. The right part shows how to connect Euclidean Shape Clues to shape geometry.
	}
        \label{fig:geoloss_shp}
        \vspace{-3mm}
\end{figure}
These displacement vectors are sufficient to represent the shape of a map instance and such a representation is invariant to translation transformations. However, we would like to note that this representation is vulnerable to rotations, which might prevent the model from learning robust geometry. To solve this problem, we propose to represent the shape with magnitudes of displacement vectors and angles between consecutive displacement vectors, as illustrated in Figure~\ref{fig:geoloss_shp}. Specifically, the $i$-th instance is denoted by $N_v$ angle values and $N_v$ magnitude values. Let $u$ be the index, the $u$-th angle value and the magnitude value of the $i$-th instance are computed as:
\begin{equation}
	\begin{aligned}
		\alpha^{i}_{u} = \langle\vv_{u}^{i}, \vv_{u+1}^{i}\rangle \,, \hspace{1.5mm}
		d^{i}_{u} = \Vert\vv_{u}^{i}\Vert_2 \,,
	\end{aligned}
\end{equation}
where $\langle \cdot \rangle$ denotes the angle between two vectors and similarly $\vv_{N_v+1}^{i} := \vv_{1}^{i}$.

The proposed representation inherently captures common geometric patterns, such as parallelism, right angles, and proper line width, by translating them into corresponding numerical patterns within this representation. For example, in Figure~\ref{fig:geoloss_shp}, a rectangle is characterized by one $90^{\circ}$ angle and equal magnitudes of opposite displacement vectors; in addition, a straight line is represented by $0^{\circ}$ angles. Beyond discussed regular cases, the Euclidean shape clues are capable of handling more complex shapes, as illustrated in scenarios (a) and (c) of Figure~\ref{fig:method} and discussed in \S~\ref{sec:vis}.

\subsubsection{Euclidean Relation Clues.} \label{sec:relation_clue}

Having highlighted the translation and rotation invariance of G-Representation and its advantages in representing the geometric shapes of individual map instances, we further introduce its ability to capture the relations between two map instances, \eg, parallelism and perpendicular, in Euclidean space.
\begin{figure}[htb]
	\centering
        \vspace{-4mm}
	\includegraphics[width=0.85\linewidth]{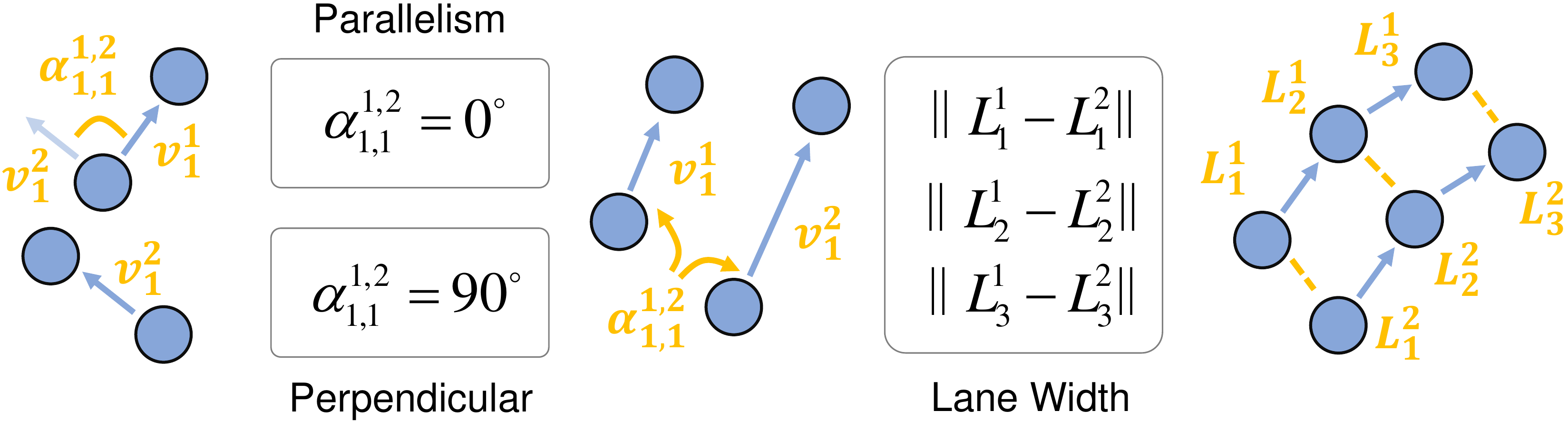}
        \vspace{-2mm}
	\caption{
		 Euclidean Relation Clues. Angles between pairs of displacement vectors on different polylines, and magnitudes of displacement vectors between point pairs indicate relation clues. Such relation clues are more superficially connected to Euclidean relation geometry as shown in the boxes.
	}
        \vspace{-3mm}
        \label{fig:geoloss_relation}
\end{figure}
Specifically, given the vanilla representations and displacement vectors of the $i$-th and $j$-th map instance, we represent the relation between instance $i$ and $j$ with 1) angles between each pair of their respective displacement vectors, and 2) magnitudes of displacement vectors between each pair of points, as illustrated in Figure~\ref{fig:geoloss_relation}. Formally, 
\begin{equation}
	\begin{aligned}
		\alpha^{i,j}_{u,v} = \langle\vv_{u}^{i},\vv_{v}^{j}\rangle \,, \hspace{1.5mm}
		d^{i,j}_{u,v} = \Vert\mL_{i,u} - \mL_{j,v}\Vert_2 \,.
	\end{aligned}
\end{equation}

This representation is also translation- and rotation-invariant. Similar to Euclidean Shape Clues, it inherently captures common relations between map instances by translating them into corresponding numerical patterns. For example, Figure~\ref{fig:geoloss_relation} shows that the perpendicular relation can be directly represented by a $90^{\circ}$ angle, and the distance between two parallel polylines, which may correspond to the width of a lane, is naturally represented by magnitudes of displacement vector between point pairs. Scenarios (a) and (b) of Figure~\ref{fig:method} show the effectiveness of Euclidean Relation Clues and more discussion can be viewed in \S~\ref{sec:vis}.

\subsection{Euclidean Loss and Objectives} \label{sec:loss}

We transform both the ground truth data and the model's predicted polylines from their original format into G-Representation. This conversion allows the model to independently optimize the angles between displacement vectors and the magnitudes of displacement vectors to more accurately align with the ground truth. The proposed Euclidean Loss is composed of two parts that measure how accurately the model predicts the shape of individual map instances and the inter-instance relations, respectively, which are denoted by $\mathcal{L}_{\rm shp}$ and $\mathcal{L}_{\rm rel}$. The Euclidean Loss can be computed as the following:
\begin{equation}
	\mathcal{L}_{\rm Euc} = \lambda_1 \cdot \mathcal{L}_{\rm shp} + \lambda_2 \cdot \mathcal{L}_{\rm rel},
\end{equation}
where $\lambda_1$ and $\lambda_2$ are hyperparameters, whose effects on the model performances are evaluated in Figure~\ref{fig:shape_rel}. Specifically,
\begin{equation}
	\begin{aligned}
        w_{i,j} &= 1 - \left( \min_{u,v} \{d_{u,v}^{i,j}\} / \sqrt{2} \right)^p \,, \hspace{2.5mm}
		\mathcal{L}_{\rm shp} = \sum_{i=1}^{N} \sum_{j=1}^{N_v} |\hat{d}^i_j - d^i_j| + \ell(\hat{\alpha}^i_j, \alpha^i_j)  \,,\\
		\mathcal{L}_{\rm rel} &= \sum_{i=1}^{N} \sum_{j>i}^{N} 
        w_{i,j} \sum_{u=1}^{N_v}\sum_{v=1}^{N_v}  \left(|\hat{d}^{i,j}_{u,v} - d^{i,j}_{u,v}| +  \ell(\hat{\alpha}^{i,j}_{u,v}, \alpha^{i,j}_{u,v}) 
        \right) \,,
	\end{aligned}
        \label{eq:loss_shp_rel}
\end{equation}
where $\ell(\cdot)$ is a function based on $\normlone$ loss. We avoid using the inverse trigonometric function to directly compute the angles, but use sine and cosine values instead:
\begin{equation}
	\ell(\hat{\alpha}, \alpha) = |\cos(\hat{\alpha}) - \cos(\alpha)| + |\sin(\hat{\alpha}) - \sin(\alpha)| \,.
\end{equation}
Moreover, the distance might influence the relation strength of instance pairs. For example, if two instances are far from each other, the relation between them might be weak. For this reason, we further adopt $w_{i,j}$ to punish weakly related instance pairs and discuss it in \S~\ref{sec:hyper_exp}. According to the best experimental results, we treat all instance pairs equally in other experiments.

Following the common practice~\cite{liao2023maptr, yuan2023streammapnet}, we also use focal loss~\cite{lin2017focal} $\mathcal{L}_{\rm cls}$ for classification. For polyline regression, we adopt point-to-point loss and edge direction loss, denoted by $\mathcal{L}_{\rm pts}$ and $\mathcal{L}_{\rm dir}$ respectively. Following~\cite{qiao2023end, liao2023maptrv2}, we further adopt segmentation loss $\mathcal{L}_{\rm seg}$ and depth estimation loss $\mathcal{L}_{\rm dep}$. These losses are detailed in appendix \S~\ref{sec:sup_loss}. The overall loss function can be written as:
\begin{equation}
	\begin{aligned}
		\mathcal{L} 
		&= \lambda\cdot \mathcal{L}_{\rm Euc} + \beta_1 \cdot \mathcal{L}_{\rm cls} + \beta_2\cdot \mathcal{L}_{\rm pts} + \beta_3\cdot \mathcal{L}_{\rm dir} + \beta_4\cdot \mathcal{L}_{\rm seg} + \beta_5\cdot \mathcal{L}_{\rm dep}.
	\end{aligned}
	\label{eq:loss_full}
\end{equation}
where hyperparameters are discussed in Figure~\ref{fig:lossw} and appendix Table~\ref{tab:sup_lhyper}.

\subsection{Geometry-Decoupled Decoder} \label{sec:dattn}
In the Transformer decoder, we obtain the $N \times N_v$ fused queries by adding $N$ instance queries to $N_v$ point queries, which is suggested in~\cite{liao2023maptr}. We denote input tokens by $\mE \in \R^{N_A\times D}$, where $D$ is the feature dimension and $N_A=N \cdot N_v$ is the number of fused queries. Intuitively, each fused query corresponds to a point on a predicted map instance. Self-Attention (SA) is formulated by:
\begin{equation}
{\rm SA}(\mE, \mM) = {\rm Softmax}\left( \frac{(\mE \mW^q)(\mE \mW^k)^\top}{\sqrt{D_k}} \odot\mM \right)\mE \mW^v \,,
\end{equation}
where $\mW^{q},\mW^{k},\mW^{v} \in \R^{D\times D_k}$ are linear projection matrices, $\mM \in \R^{N_A \times N_A}$ is the attention mask, and $\odot$ is the Hadamard product. The vanilla SA computes relations between every pair of tokens.

Nevertheless, the geometry of shape and relation pertains to distinct subsets of tokens. For any given map instance, its shape is intimately related to tokens representing that instance's points. \textit{Precise shape geometry capture requires the model to discern token correlations specific to an instance while avoiding interference from tokens of unrelated instances}. Conversely, for relation geometry modeling, it is beneficial to isolate token correlations that span across different instances, rather than those confined within a single instance.

Therefore, we present Geometry-Decoupled Attention (GDA). First, we multiply a binary mask $\mM$ to the computed attention map so that the tokens of a map instance are aggregated according to the tokens within the same map instances only, which allows the model to adjust points' positions according to learned shape geometry. Denote the index of map instance that the $i$-th token belongs to as $\mathcal{I}_i$. For example, tokens 1,2,$\dots$,$N_v$ belong to the first instance so that $\mathcal{I}_1, \mathcal{I}_2, \dots, \mathcal{I}_{N_v}=1$. Tokens $N_v$+1, $N_v$+2, $\dots$, $2\cdot N_v$ belong to the second instance so that $\mathcal{I}_{N_v+1}, \mathcal{I}_{N_v+2}, \dots, \mathcal{I}_{2N_v}=2$. With this notation, the binary mask $\mM$ can be simply constructed by:
\begin{equation}
	\mM^{\rm shp}_{i,j} = \begin{cases}
		1 &, \mathcal{I}_i = \mathcal{I}_j \\
		0 &, \mathcal{I}_i \ne \mathcal{I}_j
	\end{cases}.
\end{equation}

The second attention is expected to model relations between tokens of different map instances. The attention mask is given by:
\begin{equation}
	\mM^{\rm rel}_{i,j} = \begin{cases}
		1 &, \mathcal{I}_i \ne \mathcal{I}_j \\
		0 &, \mathcal{I}_i = \mathcal{I}_j
	\end{cases}.
\end{equation}

%% file: sec/4_experiment.tex
\section{Experiments}
\subsection{Experimental Setups}
\begin{table*}[t]
	\centering
        \vspace{-2mm}
	\caption{Comparison on the nuScenes dataset, GeMap reaches a new state-of-the-art performance. ``EB0'', ``R50'', ``PP'', ``Sec'', ``Swin-T'', and ``V2-99'' denote EfficientNet-B0~\cite{tan2019efficientnet}, ResNet50~\cite{he2016deep}, PointPillars~\cite{lang2019pointpillars}, SECOND~\cite{yan2018second}, Swin Transformer Tiny~\cite{liu2021swin}, and VoVNetV2-99~\cite{lee2019energy} respectively. Methods with two backbones utilize both camera and LiDAR inputs. ``Dns. Loss'' denotes whether any dense prediction (\eg semantic segmentation) loss is adopted. The best result is highlighted in \textbf{bold}. We reproduce all methods on a single RTX3090 GPU to test FPS for fair comparison.}
        \vspace{-3.5mm}
	\resizebox{0.94\linewidth}{!}{
		\begin{tabular}{l|cc|cccc|c}
			\toprule
			Methods & Backbone & Dns. Loss & AP$_{div}$$(\uparrow)$ & AP$_{ped}$$(\uparrow)$ & AP$_{bnd}$$(\uparrow)$ & mAP$(\uparrow)$ & FPS$(\uparrow)$\\
			\midrule
			VectorMapNet~\pub{ICML'23}~\cite{liu2023vectormapnet} & R50 &  & 42.5 & 51.4 & 44.1 & 46.0 & 5.3 \\
			MapTR~\pub{ICLR'23}~\cite{liao2023maptr} & R50 &  & 59.8 & 56.2 & 60.1 & 58.7 & \textbf{19.8} \\
                MapVR~\pub{NeurIPS'23}~\cite{zhang2024online} & R50 & & 61.8 & 55.0 & 59.4 & 58.8 & \textbf{19.8} \\
			\rowcolor{light-gray} GeMap~\pub{Ours} & R50 &  & \bfinchl{65.1}{3.3} & \bfinchl{59.8}{4.8} & \bfinchl{63.2}{3.8} & \bfinchl{62.7}{3.9} & 19.0 \\
			\midrule
			HDMapNet~\pub{ICRA'22}~\cite{li2022hdmapnet} & EB0 & \checkmark & 14.4 & 21.7 & 33.0 & 23.0 &  0.7 \\
			PivotNet~\pub{ICCV'23}~\cite{ding2023pivotnet} & R50 & \checkmark & 58.8 & 53.8 & 59.6 & 57.4 & 9.5 \\
			BeMapNet~\pub{CVPR'23}~\cite{qiao2023end} & R50 & \checkmark & 66.7 & 62.6 & 65.1 & 64.8 & 6.6 \\
			MapTRv2~\pub{Arxiv'23}~\cite{liao2023maptrv2} & R50 & \checkmark & 68.3 & \textbf{68.1} & 69.7 & 68.7 & 15.0 \\
			\rowcolor{light-gray} GeMap~\pub{Ours} & R50 & \checkmark & \bfinchl{69.8}{1.5} & 67.1 & \bfinchl{71.4}{1.7} & \bfinchl{69.4}{0.7} & \textbf{15.8}\\ \midrule
   
            HDMapNet~\pub{ICRA'22}~\cite{li2022hdmapnet} & EB0 \& PP & & 29.6 & 16.3 & 46.7 & 31.0 & 0.7 \\ 
            VectorMapNet~\pub{ICML'23}~\cite{liu2023vectormapnet} & R50 \& PP & & 60.1 & 48.2 & 53.0 & 53.7 & - \\
            MapTR~\pub{ICLR'23}~\cite{liao2023maptr} & R50 \& Sec & & 62.3 & 55.9 & 69.3 & 62.5 & 6.7 \\
            MapVR~\pub{NeurIPS'23}~\cite{zhang2024online} & R50 \& PP & & 62.7 & 60.4 & 67.2 & 63.5 & - \\
            \rowcolor{light-gray} GeMap~\pub{Ours} & R50 \& Sec & & \bfinchl{66.3}{3.6} & \bfinchl{62.2}{1.8} & \bfinchl{71.1}{3.9} & \bfinchl{66.5}{3.0} & \textbf{7.3} \\
            \midrule
            MapTRv2~\pub{Arxiv'23}~\cite{liao2023maptrv2} & R50 \& Sec & \checkmark & 65.6 & 66.5 & \textbf{74.8} & 69.0 & 6.6 \\
            \rowcolor{light-gray} GeMap~\pub{Ours} & R50 \&
            Sec & \checkmark & \bfinchl{69.8}{4.2} & \bfinchl{68.0}{1.5} & 73.4 & \bfinchl{70.4}{1.4} & \textbf{6.8} \\
            \midrule
            MapTRv2~\pub{Arxiv'23}~\cite{liao2023maptrv2} & V2-99 & \checkmark & 73.7 & 71.4 & 75.0 & 73.4 & 10.1 \\
            \rowcolor{light-gray} GeMap~\pub{Ours} & Swin-T & \checkmark & 72.8 & 70.4 & 72.8 & 72.0 & \textbf{11.4} \\
            \rowcolor{light-gray} GeMap~\pub{Ours} & V2-99 & \checkmark & \bfinchl{76.0}{2.3} & \bfinchl{74.3}{2.9} & \bfinchl{77.7}{2.7} & \bfinchl{76.0}{2.6} & 10.8 \\
			\bottomrule
		\end{tabular}
	}
	\vspace{-2mm}
	\label{tab:nus}
\end{table*}

\begin{table*}[t]
	\centering
	\caption{Following BeMapNet~\cite{qiao2023end}, we also compare GeMap on the nuScenes dataset under different weather conditions.}
        \vspace{-3.5mm}
	\resizebox{0.94\linewidth}{!}{
		\begin{tabular}{l|cc|cccc|c}
			\toprule
			Methods & Backbone & Dns. Loss & mAP$_{sun}$$(\uparrow)$ & mAP$_{cld}$$(\uparrow)$ & mAP$_{rny}$$(\uparrow)$ & mAP$_{avg}$$(\uparrow)$ & FPS$(\uparrow)$\\
			\midrule
			VectorMapNet~\pub{ICML'23}~\cite{liu2023vectormapnet} & R50 &  & 43.8 & 44.1 & 36.6 & 41.5 & 5.3 \\
			MapTR~\pub{ICLR'23}~\cite{liao2023maptr} & R50 &  & 62.1 & 60.5 & 52.8 & 58.4 & \textbf{19.8} \\
			\rowcolor{light-gray} GeMap~\pub{Ours} & R50 &  & \bfinchl{66.0}{3.9} & \bfinchl{64.3}{3.8} & \bfinchl{54.4}{1.6} & \bfinchl{61.5}{3.1} & 19.0 \\
			\midrule
			BeMapNet~\pub{CVPR'23}~\cite{qiao2023end} & R50 & \checkmark & 67.3 & 67.5 & 56.6 & 63.8 & 6.6 \\
			\rowcolor{light-gray} GeMap~\pub{Ours} & R50 & \checkmark & \bfinchl{73.1}{5.8} & \bfinchl{71.0}{3.5} & \bfinchl{59.3}{2.7} & \bfinchl{67.8}{4.0} & \textbf{15.8}\\
			\bottomrule
		\end{tabular}
	}
	\vspace{-3.5mm}
	\label{tab:nus_weather}
\end{table*}

\begin{table*}[htb]
	\centering
	\caption{Comparison on Argoverse 2 dataset. GeMap demonstrates significant performance improvements over previous methods.}
    \vspace{-3.5mm}
	\resizebox{0.94\linewidth}{!}{
		\begin{tabular}{l|cc|cccc|c}
			\toprule
			Methods & Backbone & Dns. Loss & AP$_{div}$$(\uparrow)$ & AP$_{ped}$$(\uparrow)$ & AP$_{bnd}$$(\uparrow)$ & mAP$(\uparrow)$ & FPS$(\uparrow)$\\
			\midrule
			VectorMapNet~\pub{ICML'23}~\cite{liu2023vectormapnet} & R50 &  & 36.1 & 38.3 & 39.2 & 37.9 & - \\ 
           \rowcolor{light-gray} GeMap~\pub{Ours} & R50 &  &  \bfinchl{67.6}{31.5} & \bfinchl{59.3}{21.0} & \bfinchl{64.7}{25.5} & \bfinchl{63.9}{26.0} & 16.7\\
                \midrule
                HDMapNet~\pub{ICRA'22}~\cite{li2022hdmapnet} & EB0 & \checkmark & 5.7 & 13.1 & 37.6 & 18.8 & - \\
			MapTRv2~\pub{Arxiv'23}~\cite{liao2023maptrv2} & R50 & \checkmark & 72.1 & 62.9 & 67.1 & 67.4 & 13.6\\
			\rowcolor{light-gray} GeMap~\pub{Ours} & R50 & \checkmark & \bfinchl{75.7}{3.6} & \bfinchl{69.2}{6.3} & \bfinchl{70.5}{3.4} & \bfinchl{71.8}{4.4} & \textbf{13.8}\\
			\bottomrule
		\end{tabular}
	}
	\vspace{-2mm}
	\label{tab:av2}
\end{table*}

\begin{table*}[ht]
    \centering
    \caption{Ablation study on the nuScenes dataset. We train the model for 24 epochs. * denotes replacing GDA with 2 layers of vanilla self-attention.}
    \vspace{-3.5mm}
    \resizebox{0.94\linewidth}{!}{
        \begin{tabular}{l|ccc|cccc}
        \toprule
        Method & $\mathcal{L}_{\rm shp}$ (\S~\ref{sec:shape_clue})& $\mathcal{L}_{\rm rel}$ (\S~\ref{sec:relation_clue}) & GDA (\S~\ref{sec:dattn}) & $\mathrm{AP}_{div}(\uparrow)$ & $\mathrm{AP}_{ped}(\uparrow)$ & $\mathrm{AP}_{bnd}(\uparrow)$ & $\mathrm{mAP}(\uparrow)$ \\
        \midrule
        Baseline & & & & 49.5 & 44.7 & 53.7 & 49.3 \\
        \midrule
        \rowcolor{light-gray} + Decoupled Attention & & & \checkmark & 53.4 & 46.6 & 53.5 & 51.2 \\
        \multirow{2}*{+ Single Euclidean Loss} & & \checkmark & & 51.0 & 45.4 & 52.7 & 49.7  \\
        & \checkmark & & & 51.7 & 43.4 & 53.0 & 49.4 \\
        \midrule
        \rowcolor{light-gray} + Euclidean Loss & \checkmark & \checkmark & & 51.5 & 43.9 & 51.1 & 48.8\\
        + Single Euclidean Loss	 & \checkmark & & \checkmark & 54.0 & 48.2 & 53.1 & 51.8\\
        \ \ \  with Decoupled Attention & & \checkmark & \checkmark & \textbf{54.7} & 47.3 & \textbf{55.3} & 52.4 \\
        \midrule
        Replace GDA with 2-SA & \checkmark & \checkmark & 2-SA$^{*}$ & 53.5 & 46.2 & 54.4 & 51.4 \\
        \rowcolor{light-gray} Full & \checkmark & \checkmark & \checkmark & \inchl{53.6}{4.1} & \bfinchl{49.2}{4.5} & \inchl{54.8}{1.1} & \bfinchl{52.6}{3.3} \\
        \bottomrule
        \end{tabular}
    }
    \vspace{-3mm}
    \label{tab:abl}
\end{table*}

\noindent{\textbf{Datasets.}} To evaluate GeMap, we conduct experiments on the nuScenes dataset~\cite{caesar2020nuscenes}, a widely adopted large-scale autonomous driving dataset that includes 1,000 scenes captured by six RGB cameras with a 360-degree field of view, and provides precise annotations from LiDAR point clouds for HD map construction. For comparability with prior research~\cite {li2022hdmapnet, liao2023maptr}, we focus on three static categories of map instances: pedestrian crossings, lane dividers, and road boundaries. With dataset splits provided by BeMapNet~\cite{qiao2023end}, we also evaluate GeMaps under three weather conditions: sunny, cloudy, and rainy. Additionally, we use the Argoverse 2 dataset~\cite{wilson2023argoverse} as another benchmark, which consists of approximately 108,000 frames, each providing images from seven cameras.

\noindent{\textbf{Metrics.}} We evaluate the performance of GeMap using the widely adopted metric of Average Precision (AP) \cite{li2022hdmapnet, liao2023maptr}. Specifically, we categorize a prediction as a True Positive if the Chamfer Distance between the predicted instance and its ground truth counterpart is less than a predefined threshold. For our experiments, we set these thresholds at 0.5, 1.0, and 1.5 meters.

\noindent{\textbf{Implementation Details.}} GeMap leverages 8 NVIDIA RTX 3090 GPUs for training. We adopt AdamW~\cite{loshchilov2017decoupled} as the optimizer and utilize Cosine Annealing with a linear warm-up phase~\cite{loshchilov2016sgdr} as the learning rate scheduler. For more details on hyperparameters, please refer to appendix \S~\ref{sec:sup_hyperS}.

\subsection{Main Results}

\noindent\textbf{Results on nuScenes.} As delineated in Table~\ref{tab:nus}, GeMap delivers a state-of-the-art performance, achieving a mean Average Precision (mAP) of 69.4\% using the camera-only setup and ResNet50~\cite{he2016deep} backbone. GeMap particularly improves the precision for identifying dividers and boundaries, with enhancements of +1.5\% and +1.7\%, respectively. Importantly, these performance gains do not come at a high cost of efficiency, as GeMap sustains inference speeds on par with and even exceeds the previously established state-of-the-art, measured in frames per second (FPS). With more powerful vision backbones, GeMap can achieve significantly improved performance. Specifically, VoVNetV2-99~\cite{lee2019energy} and Swin-T~\cite{liu2021swin} outperform the ResNet50 baseline by $+6.6\%$ and $+2.6\%$ in ${\rm mAP}$, respectively. Also, GeMap achieves significantly improved performance with extra LiDAR inputs, which indicates the generalizability of GeMap to multi-modality settings. Moreover, we also compare performance under different weather conditions in Table~\ref{tab:nus_weather}. GeMap exceeds previous works under sunny, cloudy, and rainy scenarios, which shows the potential of geometry to raise model robustness to varied weather conditions. 

\noindent\textbf{Results on Argoverse 2.} Extending our evaluation to the Argoverse 2 dataset, as presented in Table~\ref{tab:av2}, GeMap also presents a new SOTA performance of 71.8\% mAP, outperforming the highly advanced MapTRv2 model by +4.4\%. This achievement not only demonstrates the effectiveness of GeMap on a different dataset but also emphasizes the adaptability and precision of GeMap in the evolving landscape of autonomous driving technologies.

\subsection{Ablation Study}

Ablation studies on the nuScenes dataset are carried out to evaluate the individual contributions of GeMap's components and other vision backbones.

\noindent\textbf{Components Ablation.} The ablation experiments, detailed in Table~\ref{tab:abl}, involved training the model for a reduced duration of 24 epochs without dense prediction losses, to facilitate efficient analysis. These experiments confirm the significant role of GDA, which alone increases mAP by +1.9\%. Moreover, applying the full suite of components results in a further mAP enhancement of +1.4\%. A notable discovery is the detrimental effect on model performance when Euclidean Loss is applied without GDA, which leads to a 0.5\% reduction in mAP. This reinforces our position that conventional self-attention mechanisms are insufficient for encoding a variety of geometric properties.

\begin{figure*}[t]
	\centering
        \vspace{-3mm}
	\begin{minipage}{0.98\linewidth}
		\begin{center}
			\includegraphics[width=1.0\linewidth]{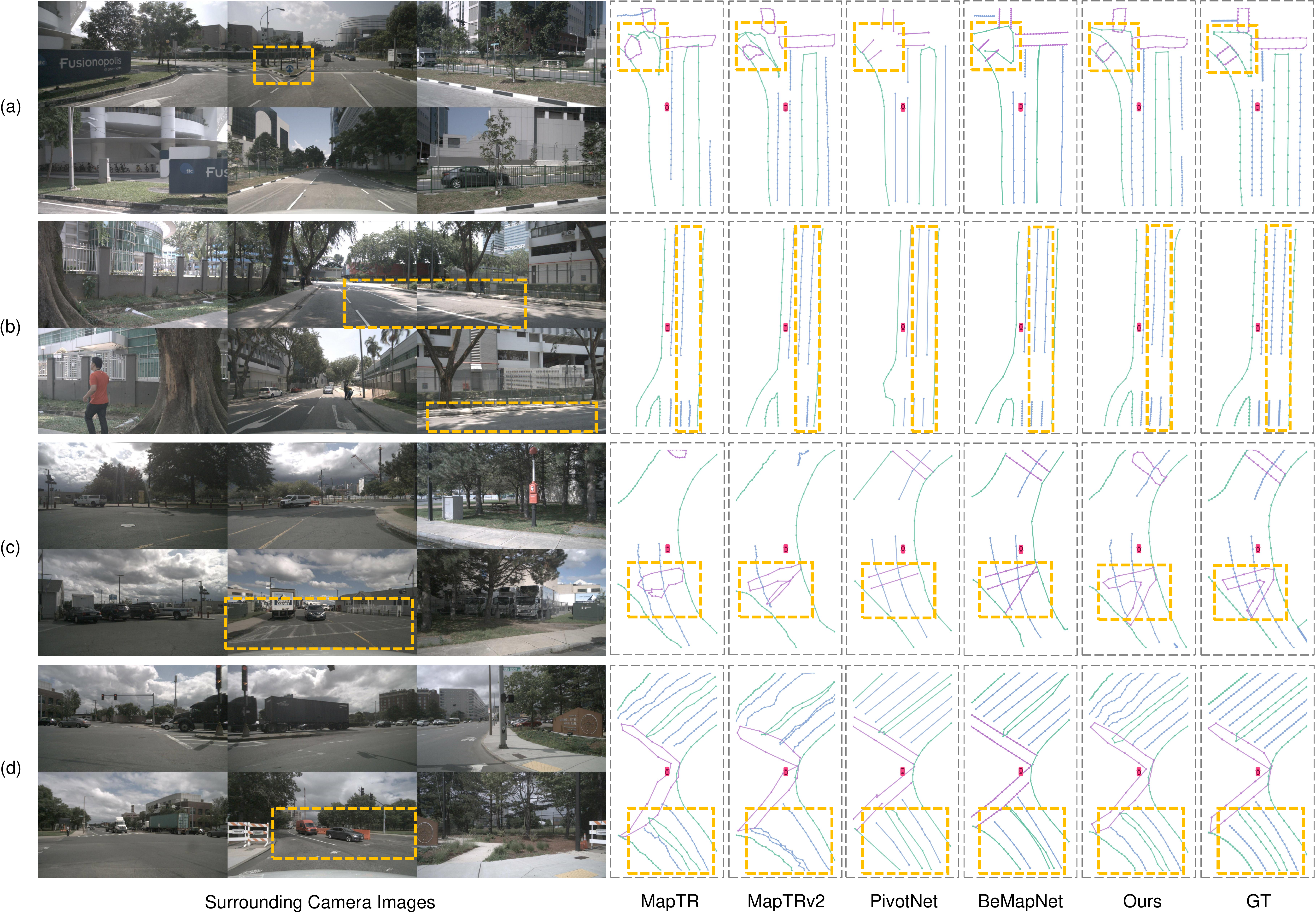}
		\end{center}
	\end{minipage}
        \vspace{-2mm}
	\caption{
		Visualization results. Instances that are hard to construct are highlighted in orange boxes. Scenario (a) depicts a complex triangular road boundary. Scenario (b) includes a divider that is hard to recognize according to strong sunlight. Scenario (c) depicts pedestrian crossings that can only be partially observed. In scenario (d), the BEV map is tilted and lane markings are obscured by vehicles. These challenging cases indicate the superiority and robustness of GeMap.
	}
        \vspace{-3mm}
	\label{fig:method}
 \vspace{-3mm}
\end{figure*}

\noindent\textbf{Effectiveness of GDA.} In the proposed GDA (\ref{sec:dattn}), we decouple shape and relation learning, sequentially applying Euclidean shape and relation attention. In addition, we also attempt to intuitively double the self-attention layers in one block, referred to as ``2-SA''. As shown in Table~\ref{tab:abl}, GDA outperforms vanilla self-attention by +1.2\% mAP, indicating the superiority of GDA when combined with Euclidean Loss.

\begin{figure}[t]
	\centering
	\subfloat[
	\label{fig:shape_rel}
	]{
		\begin{minipage}{0.44\linewidth}{\begin{center}
					{\includegraphics[width=1\linewidth]{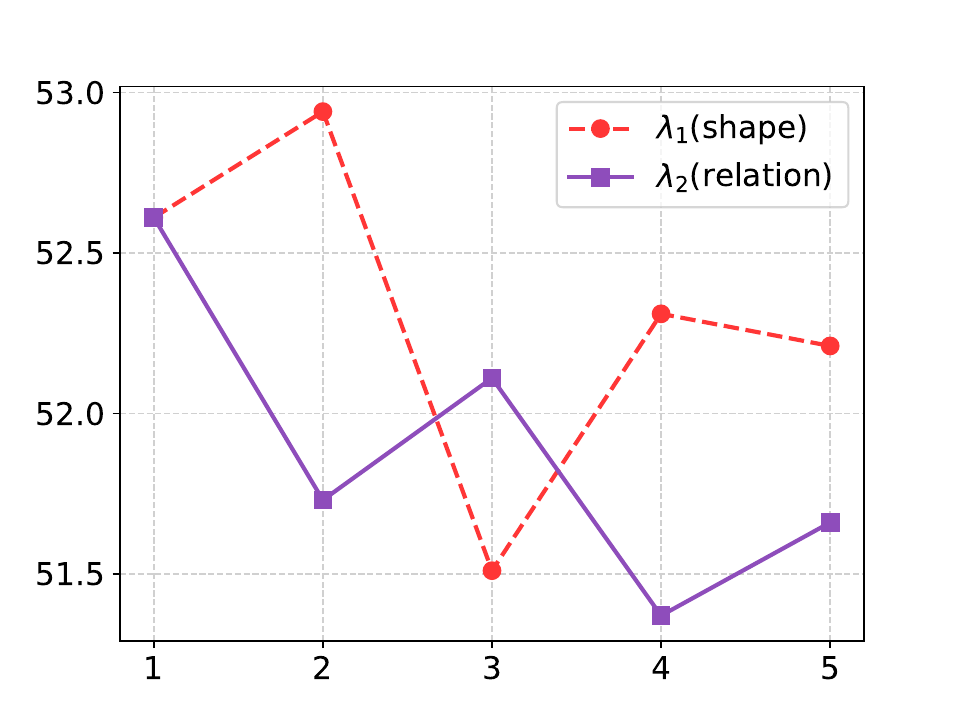}}
		\end{center}}\end{minipage}
	}
	\subfloat[
	\label{fig:lossw}
	]{
		\begin{minipage}{0.44\linewidth}{\begin{center}
					{\includegraphics[width=1\linewidth]{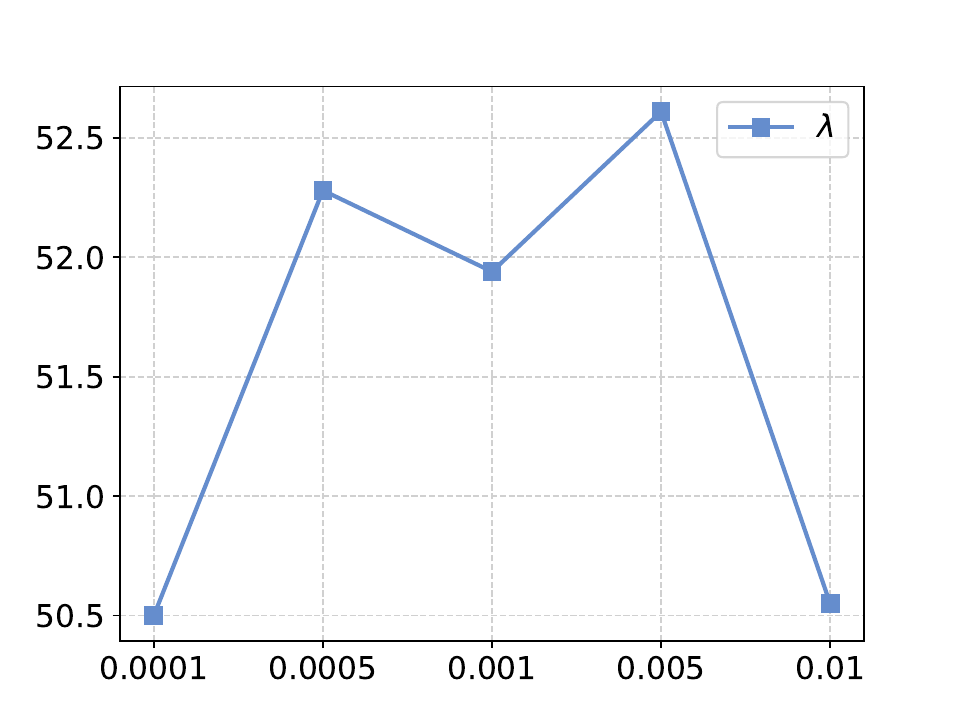}}
		\end{center}}\end{minipage}
	}
        \vspace{-3mm}
	\caption{
		Hyperparameter experiments. (a) The best performance is reached with balanced Euclidean shape loss and relation loss. (b) Model performances degrade with too large or too small $\lambda$.
	}
        \vspace{-6mm}
	\label{fig:hyper}
\end{figure}

\subsection{Visualization Analysis} \label{sec:vis}

\noindent\textbf{More Precise Shape Awareness.} Scenario (a) in Figure~\ref{fig:method} demonstrates GeMap's capability in accurately identifying complex road boundary shapes, such as triangular boundaries. This precise result is attributed to the shape geometry we propose, which allows for an inherent description of the triangle feature, in contrast to baselines which do not discern it as effectively. A more irregularly shaped pedestrian crossing can be viewed in scenario (c) of Figure~\ref{fig:method} and GeMap provides more precise construction than other methods.

\noindent\textbf{Better Relation Awareness.} With the help of Euclidean Relation Clues, GeMap can better infer parallelism and perpendicular in scenario (a) of Figure~\ref{fig:method}. More interestingly, in scenario (b), the strong sunlight makes the divider highlighted hard to recognize. However, according to the understanding of lane width patterns, GeMap has the sense that there should be a divider.

\noindent\textbf{Alleviating Occlusion Issues.} The scenarios depicted (c) of Figure~\ref{fig:method} demonstrate the utility of our shape geometry in alleviating challenges associated with occlusions in partially visible instances. In particular, scenario (c) features a black car that obscures part of a pedestrian crossing; nonetheless, GeMap successfully deduces the overall structure of the crossing within a complex shape. In contrast, these baseline models~\cite{liao2023maptr,liao2023maptrv2,ding2023pivotnet} struggle with such complex shape recovery under similar occlusion conditions.

\noindent\textbf{Enhanced Robustness to Rotational Transformations.} Scenario (d) in Figure~\ref{fig:method} exemplifies the resilience of GeMap to rotational transformations, as evidenced when the ego-vehicle executes a right turn causing the BEV map to appear highly tilted. Additionally, this scenario features lanes that are extensively obscured by vehicular traffic. Despite these challenges, GeMap more adeptly maintains the integrity of lane width and parallelism, which are key geometry, in contrast to the baseline model. This underscores the superior robustness of our geometric constructs against rotational distortions.

\subsection{Hyperparameter Experiments} \label{sec:hyper_exp}

\noindent\textbf{Comparative Impact of Euclidean Losses.} Figure~\ref{fig:shape_rel} shows that GeMap exhibits larger sensitivity to the shape loss than to the relation loss, as evidenced by the steeper change of the performance curve. The optimal result is obtained when the shape and relation losses are relatively balanced by weights, implying that both contribute comparably to the model's optimization process.

\noindent\textbf{Optimization of Euclidean Loss Weighting.} The performance of the model, as demonstrated in Figure~\ref{fig:lossw}, peaks when the weight of the Euclidean Loss, $\lambda$, is set to $5\times {10}^{-3}$
 . Notably, model performance degrades with a too large or too small $\lambda$, which informed us to use a $\lambda$ value of $5\times 10^{-3}$ for our experiments.

\noindent\textbf{Distance Weighting.} Experimental results of distance weighting are presented in Table~\ref{tab:dw}. We change the order of distance ($p$) in Equation~\ref{eq:loss_shp_rel} and $p\in \{1, 2, 4\}$ corresponds to ``Linear'', ``Square'' and ``4th Power'' respectively. Moreover, we also try to treat instance pairs equally and it delivers the best performance.
\begin{table}[htb]
    \centering
    \vspace{-3mm}
    \caption{Distance weighting on the nuScenes dataset.}
    \vspace{-3.5mm}
    \resizebox{0.5\linewidth}{!}{
        \begin{tabular}{l|cccc}
            \toprule
            Strategy & $\text{AP}_{div}(\uparrow)$ & $\text{AP}_{ped}(\uparrow)$ & $\text{AP}_{bnd}(\uparrow)$ & mAP$(\uparrow)$ \\
            \midrule
            Linear & 53.5 & 46.6 & \textbf{56.2} & 52.1\\
            Square & 53.2 & 47.9 & 55.2 & 52.1 \\
            4th Power & 53.2 & 47.4 & 55.8 & 52.1 \\
            Equal & \textbf{53.6} & \textbf{49.2} & 54.8 & \textbf{52.6} \\
            \bottomrule
        \end{tabular}
    }
    \vspace{-6mm}
    \label{tab:dw}
\end{table}

%% file: sec/5_conclusion.tex
\section{Conclusion}
\vspace{-1.5mm}
In this paper, we realize significant shape and relation geometry inherent in HD map instances and propose the GeMap. GeMap includes the integration of Euclidean shape and relation losses for auxiliary supervision. To further refine the model's awareness of diverse geometry, we introduce the Geometry-Decoupled Attention mechanism. GeMap has achieved state-of-the-art performances on both the nuScenes and Argoverse 2 datasets, underscoring its effectiveness. Despite these promising results, the current application of geometry remains fundamental, and future research could focus on more sophisticated representations or enhanced geometric patterns. Furthermore, the application of geometry extends beyond HD map construction, offering potential solutions to occlusion challenges in other autonomous driving tasks. We anticipate that these findings will inspire further research.

%% file: sec/X_suppl.tex
\clearpage
\appendix
\setcounter{table}{0}
\setcounter{figure}{0}
\renewcommand{\thetable}{A\arabic{table}}
\renewcommand{\thefigure}{A\arabic{figure}}
\setcounter{page}{1}
{
\section*{\centering{Online Vectorized HD Map Construction using Geometry \\
\textit{Supplementary Material} }}
}

This supplementary material is organized as follows:
\begin{itemize}
	\item More details on the method design (\S~\ref{sec:sup_details}).
	\item Further quantitative experimental results (\S~\ref{sec:sup_exp}).
	\item Additional visualization results under three weather conditions (\S~\ref{sec:sup_vis}).
\end{itemize}

\section{Additional Details} \label{sec:sup_details}
\subsection{Objective Functions} \label{sec:sup_loss}
\noindent\textbf{Objective Configurations.} Our method employs two distinct objective functions. The full objective function is defined as follows:
\begin{equation}
	\begin{aligned}
		\mathcal{L} 
		&= \lambda\cdot \mathcal{L}_{\rm Euc} + \beta_1 \cdot \mathcal{L}_{\rm cls} + \beta_2\cdot \mathcal{L}_{\rm pts} \\
		&+ \beta_3\cdot \mathcal{L}_{\rm dir} + \beta_4\cdot \mathcal{L}_{\rm seg} + \beta_5\cdot \mathcal{L}_{\rm dep}
	\end{aligned}
	\label{eq:sup_loss_full}
\end{equation}
and the simpler one which excludes dense prediction losses is:
\begin{equation}
	\begin{aligned}
		\mathcal{L}' = \lambda\cdot \mathcal{L}_{\rm Euc} + \beta_1 \cdot \mathcal{L}_{\rm cls} + \beta_2\cdot \mathcal{L}_{\rm pts}  
		+ \beta_3\cdot \mathcal{L}_{\rm dir}.
	\end{aligned}
	\label{eq:sup_loss_simple}
\end{equation}
\noindent\textbf{Point Order Agnostic Matching.} In accordance with the methodology proposed by MapTR~\cite{liao2023maptr}, we employ point order-agnostic matching between the prediction and ground truth. In the subsequent formulations, we assume that the prediction and ground truth have already been paired.

\noindent\textbf{Classification Loss.} To enhance the model's comprehension of semantics associated with various map instance types, we incorporate the classification task. Let $\hat{\vp} \in \R^{N\times C}$ denote the predicted probabilities, where $C$ is the number of instance categories. Here, $\hat{\vp}_{ic}$ represents the predicted probability of instance $i$ belonging to category $c$. With ground truth labels $\vy \in \{1,...,C\}^{N}$, the objective function based on focal loss is defined as follows:
\begin{equation}
	\mathcal{L}_{\rm cls} = -\sum_{i=1}^{N} \sum_{c=1}^{C} \delta[\vy_i=c]\cdot \alpha_c(1-\hat{\vp}_{ic})^\gamma\log \hat{\vp}_{ic} \,,
\end{equation}
where $\delta[q] = 1$ if proposition $q$ is true and $\delta[q] = 0$ otherwise.

\noindent\textbf{Point Loss.} For the perception of instance positions, we employ a point loss that evaluates $\normlone$ distances between predicted points and ground truth points, which is specified as:
\begin{equation}
	\mathcal{L}_{\rm pts} = \sum_{i=1}^{N} \sum_{j=1}^{N_v} \Vert \hat{\mL}_{j}^{i} - \mL _{j}^{i}\Vert_1.
\end{equation}

\noindent\textbf{Edge Direction Loss.} To obtain more precise displacement vectors, which are crucial in our G-Representation, we incorporate an edge direction loss. This loss quantifies the cosine similarity between predicted displacement vectors and their corresponding ground truth vectors. Specifically, the loss is defined as:
\begin{equation}
	\mathcal{L}_{\rm dir} = -\sum_{i=1}^{N} \sum_{j=1}^{N_v} \frac{(\hat{\vv}_{j}^{i})^{\top} \vv_{j}^{i}}{\Vert \hat{\vv}_{j}^{i}\Vert_2\cdot\Vert \vv_{j}^{i}\Vert_2}.
\end{equation}

\noindent\textbf{Segmentation Loss.} The auxiliary binary segmentation task is valuable for assisting the model in the coarse perception of shape geometry. We integrate a convolutional neural network-based BEV segmentation head with BEV features. Let $\hat{\mP}_{\rm bev} \in \R^{H'\times W'}$ represent the probability of each grid belonging to the instance area, and $\mY_{\rm bev} \in \{0, 1\}^{H'\times W'}$ denote the ground truth. The corresponding objective function is defined as:
\begin{equation}
	\mathcal{L}_{\rm bev} = \mathcal{L}_{\rm bce}(\hat{\mP}_{\rm bev}, \mY_{\rm bev}) \,, \label{eq:sup_seg_loss}
\end{equation}
where the binary cross entropy loss $\mathcal{L}_{\rm bce}$ is:
\begin{equation}
	\begin{aligned}
		\mathcal{L}_{\rm bce}(\hat{p}, y) 
		= &-\delta[y=1]\cdot \log \hat{p} \\
		&- \delta[y=0]\cdot \log (1-\hat{p}).
	\end{aligned}
\end{equation}
We also introduce the auxiliary PV segmentation task, incorporating a shared convolutional neural network head for all views. The ground truth is projected back to the PV space to form the binary mask. Let $\hat{\mP}_{\rm pv}^{k}\in \R^{H \times W}$ denote the segmentation results for view $k$ with corresponding ground truth $\mY_{\rm pv}^k\in \{0, 1\}^{H\times W}$, then the objective function can be expressed as:
\begin{equation}
	\mathcal{L}_{\rm pv} = \sum_{k=1}^{K} \mathcal{L}_{\rm bce} (\hat{\mP}_{\rm pv}^k, \mY_{\rm pv}^k).
\end{equation}
Finally, we obtain the segmentation loss as follows:
\begin{equation}
	\mathcal{L}_{\rm seg} = \beta_{\rm bev} \cdot \mathcal{L}_{\rm bev} + \beta_{\rm pv} \cdot \mathcal{L}_{\rm {pv}}.
\end{equation}

\noindent\textbf{Depth Estimation Loss.} To enhance depth perception, we adopt an auxiliary depth estimation task.
Let $\hat{\mP}^{k}_{\rm dep} \in \R^{H \times W \times D}$ represent the depth distribution of each grid estimated by LSS~\cite{philion2020lift} in the PV space of view $k$, where $D$ represents the number of quantified depth buckets. Given the ground truth $\mY_{\rm dep}^k \in \{1, ..., D\}^{H\times W\times D}$, the depth estimation loss is defined as:
\begin{equation}
	\mathcal{L}_{\rm dep} = -\sum_{k=1}^{K} \sum_{d=1}^{D}\delta[\mY_{\rm dep}^k=d] \cdot \log \hat{\mP}_{\rm dep}^k.
    \label{eq:sup_dep_loss}
\end{equation}

\subsection{Hyperparameter Settings} \label{sec:sup_hyperS}
In the default optimization setting, we set the dropout rate to $0.1$ and weight decay to $0.03$. The first 500 iterations involve a linear warm-up, starting from $1/3$ of the maximum learning rate. In the Cosine Annealing scheduler, the minimum learning rate is set to $0.001$ of the maximum. Unless explicitly stated otherwise, we train our model for 110 epochs on nuScenes and 24 epochs on Argoverse 2.
For the simplified objective configuration, we set the maximum learning rate to $6\times 10^{-4}$ with a batch size of 4. When LiDAR input is utilized, the batch size is reduced to 3. In the full objective configuration, varied hyperparameters are detailed in Table~\ref{tab:sup_hyper}. Also, the default hyperparameter settings for objective functions are presented in Table~\ref{tab:sup_lhyper}.
\begin{table}[htb]
\centering
\caption{Hyperparameters of objective functions.}
    \vspace{-2.5mm}
\resizebox{0.45\linewidth}{!}{
    \begin{tabular}{l|ccccc}
        \toprule
        Parameter & $\alpha_c$ & $\gamma$ & $\lambda$ & $\beta_1$ & $\beta_2$ \\
        \midrule
        Value & 0.25 & 2 & 0.005 & 2 & 5 \\
        \midrule
        \midrule
        Parameter & $\beta_3$ & $\beta_4$ & $\beta_5$ & $\beta_{\rm bev}$ & $\beta_{\rm pv}$ \\
        \midrule
        Value & 0.005 & 1 & 3 & 1 & 2 \\ 
        \bottomrule
    \end{tabular}
}
    \vspace{-10mm}
    \label{tab:sup_lhyper}
\end{table}
\begin{table}[htb]
	\centering
	\caption{Hyperparameters under different vision backbones.}
        \vspace{-2.5mm}
	\resizebox{0.5\linewidth}{!}{
		\begin{tabular}{l|cc}
				\toprule
				Backbone & Max Learning Rate & Batchsize \\
				\midrule
				R50 & $6\times10^{-4}$ & 4 \\
				V2-99 & $6\times10^{-4}$ & 3 \\
				Swin-T & $4\times10^{-4}$ & 3 \\
				\bottomrule
		\end{tabular}
	}
	\label{tab:sup_hyper}
\end{table}

Moreover, we set the number of instance queries as $N = 50$ and the number of point queries as $N_v = 20$. We employ a single layer of encoder in GKT and incorporate 6 attention blocks in the Geometry-Decoupled Decoder. In the context of LSS transformation, the depth spans from 1 to 35 meters, quantified at intervals of 0.5 meters, resulting in $D=68$.

\section{More Experimental Results} \label{sec:sup_exp}
In this section, we present additional ablation studies and hyperparameter experiment results. In all of these experiments, the model is trained for 24 epochs on nuScenes using the simplified objective function. Unless otherwise specified, we employ the default settings outlined in \S~\ref{sec:sup_hyperS}.

\subsection{Impact of the Decoder Block Number} \label{sec:sup_block}
We evaluate the impact of decoder block numbers on the model performance, as presented in Table~\ref{tab:sup_dblock}. When increasing the number of blocks from 1 to 6, the mAP increases by $+20.8\%$. However, naively adding more blocks might be detrimental to model performance. For example, mAP decreases by $-4.7\%$ when increasing the number of blocks from 6 to 12.
\begin{table}[htb]
	\centering
	\caption{Impact of the decoder block number. The default setting utilized in our experiments is highlighted in gray.}
        \vspace{-2.5mm}
	\resizebox{0.5\linewidth}{!}{
		\begin{tabular}{l|cccc}
			\toprule
			\# Block & AP$_{div}$$(\uparrow)$ & AP$_{ped}$$(\uparrow)$ & AP$_{bnd}$$(\uparrow)$ & mAP$(\uparrow)$\\
			\midrule
			1 & 33.5 & 24.7 & 37.3 & 31.8\\
			2 & 42.1 & 38.9 & 48.2 & 43.1\\
			4 & 51.1 & 43.5 & 53.9 & 49.5\\
			\rowcolor{light-gray} 6 & 53.6 & \textbf{49.2} & \textbf{54.8} & \textbf{52.6} \\
			8 & \textbf{54.5} & 46.4 & 53.4 & 51.4\\
			10 & 52.4 & 45.7 & 53.5 & 50.5 \\
			12 & 49.6 & 45.1 & 48.9 & 47.9 \\
			\bottomrule
		\end{tabular}
	}
	\label{tab:sup_dblock}
\end{table}

\subsection{Impact of the Query Number}
We also evaluate the influence of query numbers on model performance, as detailed in Table~\ref{tab:sup_n} for instance queries and Table~\ref{tab:sup_nv} for point queries. 

\noindent\textbf{Instance Queries.} As depicted in Table~\ref{tab:sup_n}, augmenting the number of instance queries could be advantageous for the model's performance. More specifically, the mAP exhibits an increment of $+27.8\%$ when the query number is elevated from 10 to 50. This observation aligns with intuition, as a higher number of instance queries implies a broader pool of diverse candidates.

\noindent\textbf{Point Queries.} It is observed from Table~\ref{tab:sup_nv} that an excess or insufficient number of point queries has an adverse impact on the model performance. Notably, an interesting finding is that the optimal query number varies according to different instance categories. For example, lane dividers exhibit better performance with $N_v = 10$, while pedestrian crossings and road boundaries show optimal results with $N_v = 20$. This discrepancy is attributed to the straight shape of lane dividers, whereas pedestrian crossings and road boundaries, characterized by more intricate shapes, benefit from a relatively larger point query number. Hence, the results suggest that adapting point query numbers based on the complexity of instance geometry could further enhance the model performance, which is a topic left for future investigation.
\begin{table}[htb]
	\centering
	\caption{Impact of the instance query number.}
        \vspace{-2.5mm}
	\resizebox{0.5\linewidth}{!}{
		\begin{tabular}{l|cccc}
			\toprule
			$N$ & AP$_{div}$$(\uparrow)$ & AP$_{ped}$$(\uparrow)$ & AP$_{bnd}$$(\uparrow)$ & mAP$(\uparrow)$ \\
			\midrule
			10 & 30.2 & 12.3 & 31.9 & 24.8 \\
                30 & 50.6 & 43.4 & 50.5 & 48.2 \\
			40 & 51.0 & 47.5 & 53.1 & 50.5 \\
			\rowcolor{light-gray} 50 & \textbf{53.6} & \textbf{49.2} & 54.8 & \textbf{52.6} \\
			60 & 52.6 & 49.0 & \textbf{55.6} & 52.4 \\
			\bottomrule
		\end{tabular}
	}
	\label{tab:sup_n}
        \vspace{-10mm}
\end{table}
\begin{table}[htb]
	\centering
	\caption{Impact of the point query number.}
        \vspace{-2.5mm}
	\resizebox{0.5\linewidth}{!}{
		\begin{tabular}{l|cccc}
			\toprule
			$N_v$ & AP$_{div}$$(\uparrow)$ & AP$_{ped}$$(\uparrow)$ & AP$_{bnd}$$(\uparrow)$ & mAP$(\uparrow)$ \\
			\midrule
			5 & 49.7 & 31.4 & 41.8 & 41.0 \\
			10 & \textbf{53.7} & 45.9 & 52.5 & 50.7 \\
			\rowcolor{light-gray} 20 & 53.6 & \textbf{49.2} & \textbf{54.8} & \textbf{52.6} \\
			30 & 50.9 & 48.3 & 54.7 & 51.3 \\
			40 & 50.2 & 47.9 & 54.6 & 50.9 \\
			\bottomrule
		\end{tabular}
	}
	\label{tab:sup_nv}
\end{table}

\section{More Visualization Results} \label{sec:sup_vis}
\begin{figure*}[htb]
	\centering
	\begin{minipage}{0.98\linewidth}
		\begin{center}
			\includegraphics[width=1.0\linewidth]{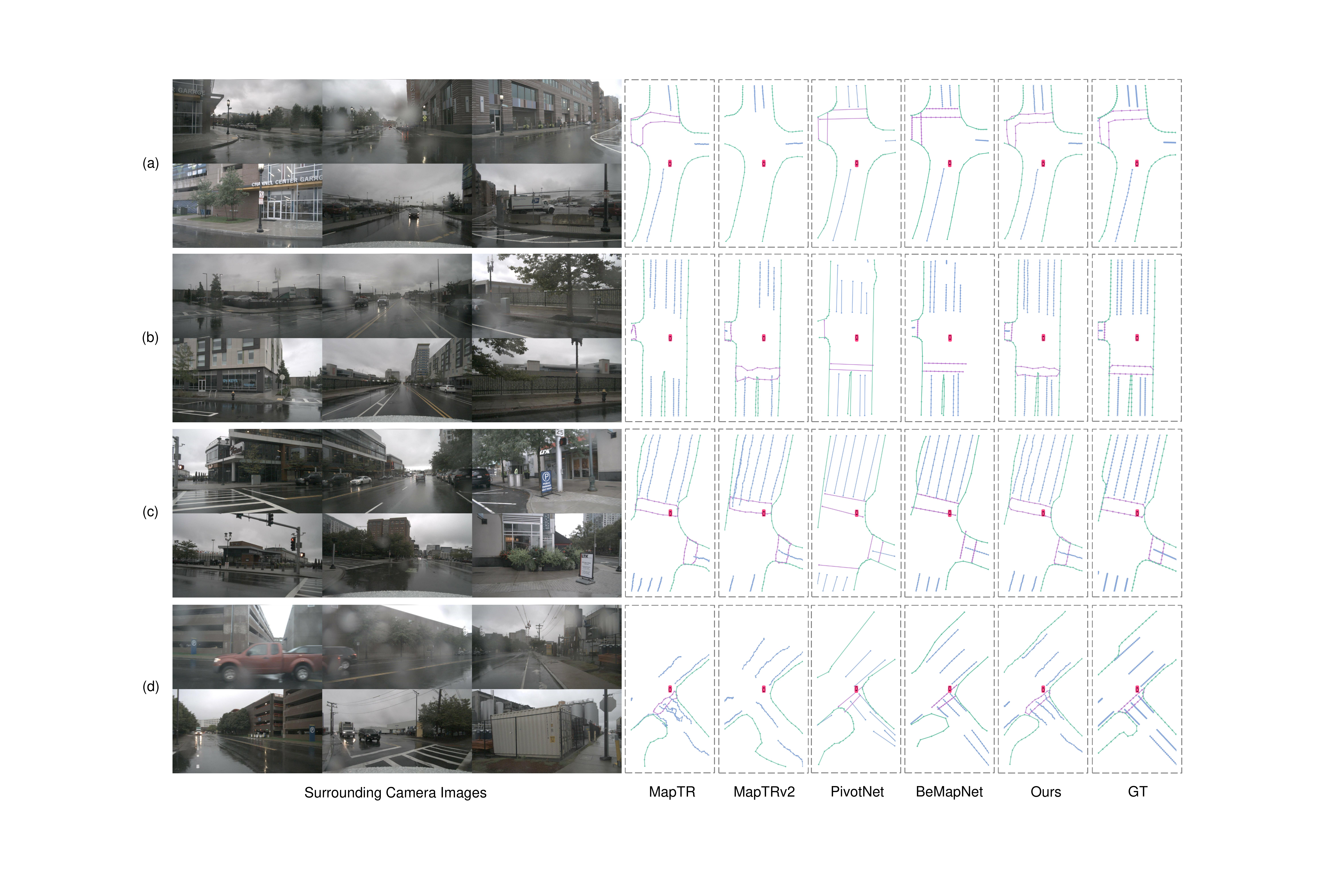}
		\end{center}
	\end{minipage}
	\caption{
		Visualization results under challenging rainy weather conditions. Even with noisy reflections on the road and map instances occluded by water drops, our method still provides robust predictions.
	}
	\label{fig:sup_rainy}
\end{figure*}

\begin{figure*}[htb]
	\centering
	\begin{minipage}{0.98\linewidth}
		\begin{center}
			\includegraphics[width=1.0\linewidth]{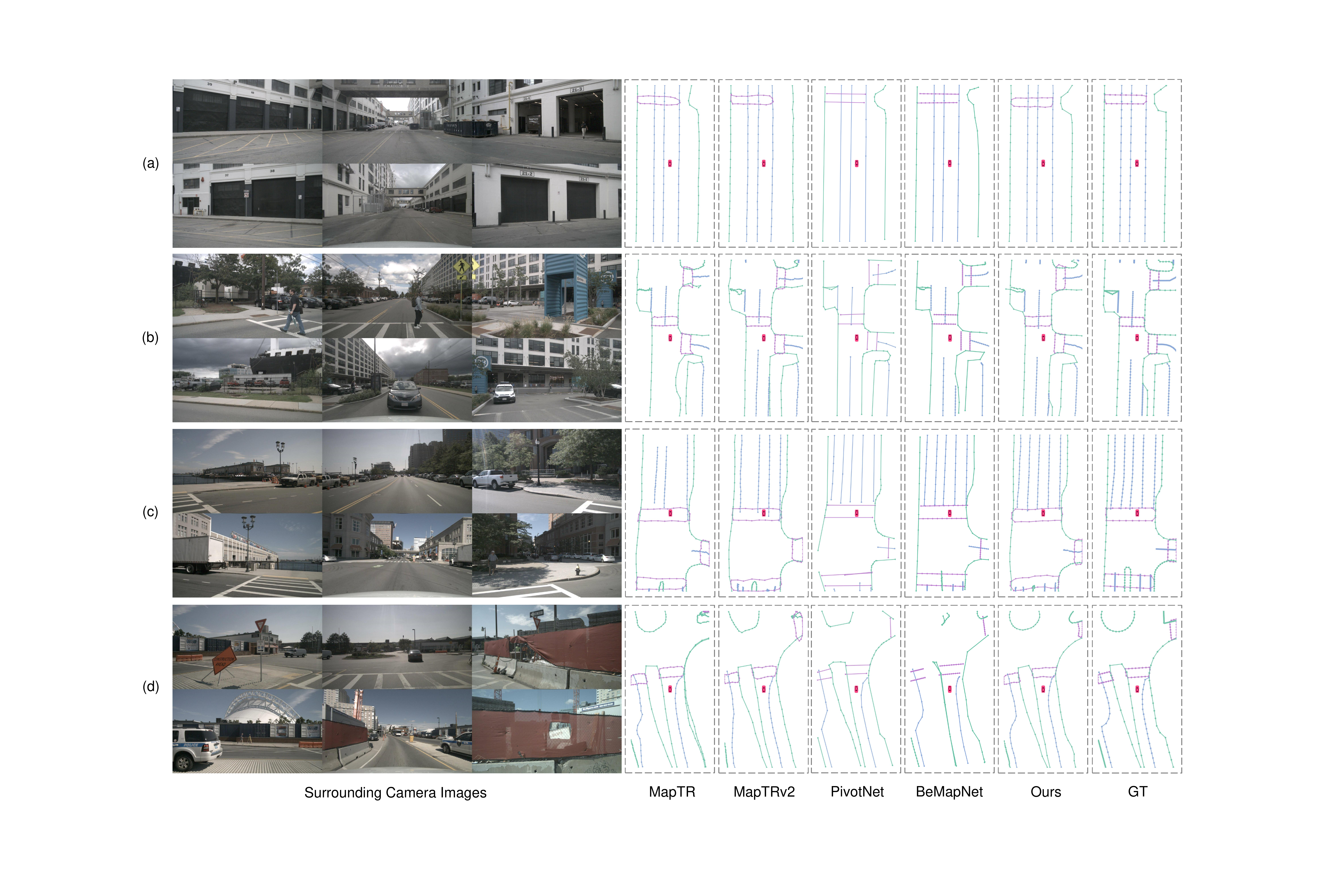}
		\end{center}
	\end{minipage}
	\caption{
		Visualization results under sunny weather conditions.
	}
	\label{fig:sup_cloudy}
\end{figure*}

\begin{figure*}[htb]
	\centering
	\begin{minipage}{0.98\linewidth}
		\begin{center}
			\includegraphics[width=1.0\linewidth]{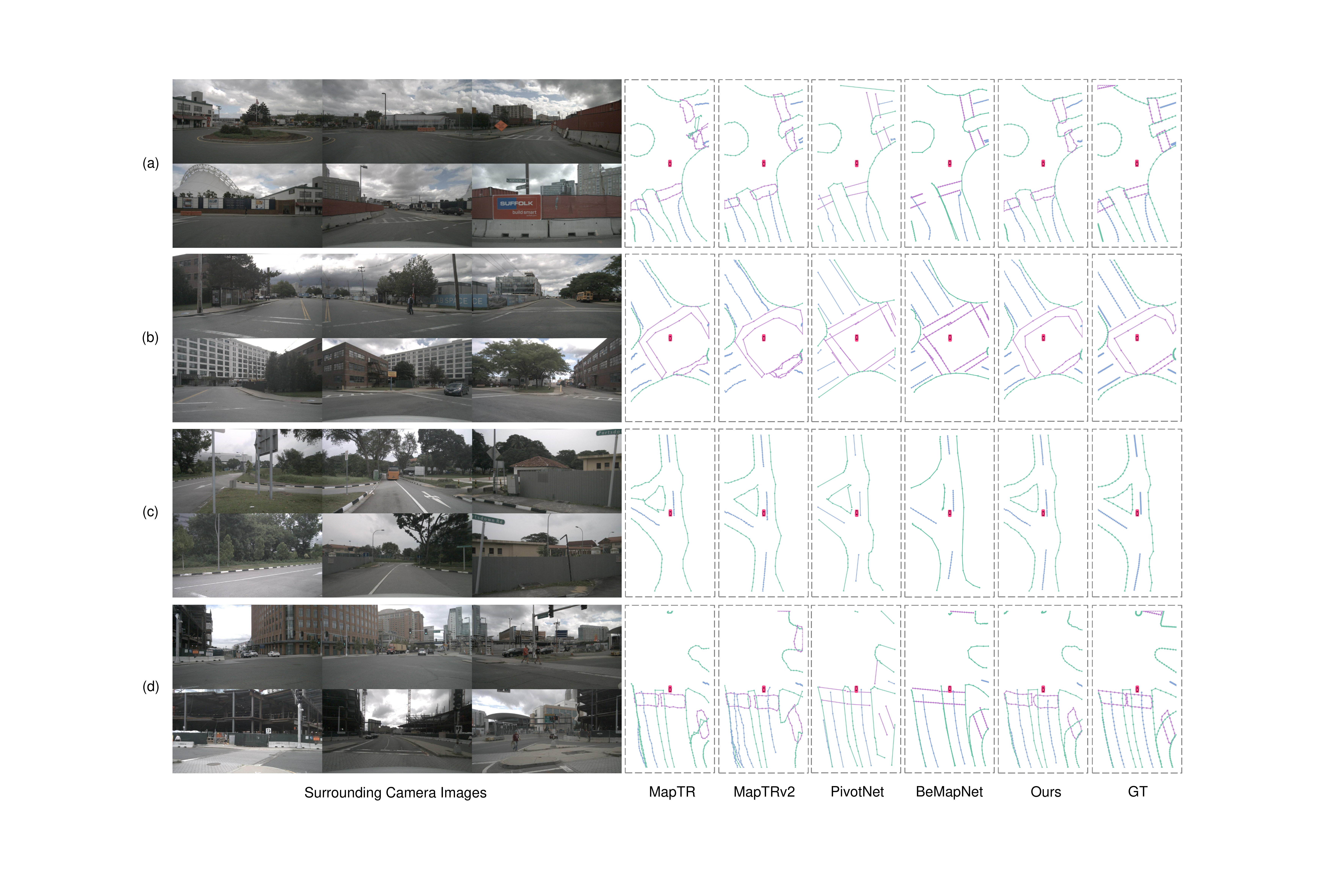}
		\end{center}
	\end{minipage}
	\caption{
		Visualization results under cloudy weather conditions.
	}
	\label{fig:sup_sunny}
\end{figure*}

We present additional visualization cases under varied weather conditions, as illustrated in Figure~\ref{fig:sup_rainy} to Figure~\ref{fig:sup_sunny}. Our method is trained with a ResNet50 backbone using the simplified objective function.

As illustrated in Figure~\ref{fig:sup_rainy}, in challenging rainy conditions, our method demonstrates more robust results. Particularly in scenario (d) of Figure~\ref{fig:sup_rainy}, where the front road boundary and lane divider are heavily occluded by water on the front windshield, our method can still recover the entire instance accurately from observed parts. This showcases the potential of proposed geometric designs.
